\documentclass[letterpaper, 10pt]{IEEEtran}
\IEEEoverridecommandlockouts
   \usepackage[pdftex]{graphicx}
 
  \graphicspath{{../pdf/}{../jpeg/}}
  \DeclareGraphicsExtensions{.pdf,.jpeg,.png}
  
  \usepackage{amssymb} 
  \usepackage{graphicx}
\usepackage[natural]{xcolor}

\usepackage{changes}

\usepackage{siunitx}

\usepackage[cmex10]{amsmath}
\usepackage[tight,footnotesize]{subfigure}

\usepackage{url}

\begin{document}

\title{\LARGE \bf 
	Cooperative Starting Movement Detection of Cyclists Using \\
	 Convolutional Neural Networks and a Boosted Stacking Ensemble}

\author{Maarten Bieshaar, Stefan Zernetsch, Andreas Hubert, Bernhard Sick, and Konrad Doll
	\thanks{M. Bieshaar, and B. Sick are with the Intelligent Embedded Systems Lab, University of Kassel,
		Kassel, Germany
		{\tt\footnotesize mbieshaar@uni-kassel.de, bsick@uni-kassel.de}}
	\thanks{S. Zernetsch, A. Hubert and K. Doll are with the Faculty of Engineering,
		University of Applied Sciences Aschaffenburg, Aschaffenburg, Germany
		{\tt\footnotesize stefan.zernetsch@h-ab.de, andreas.hubert@h-ab.de, konrad.doll@h-ab.de}}
}

\maketitle

\begin{abstract}
In future, vehicles and other traffic participants will be interconnected and equipped with various types of sensors, allowing for cooperation on different levels, such as situation prediction or intention detection.
In this article we present a cooperative approach for starting movement  detection of cyclists using 
a boosted stacking ensemble approach realizing feature- and decision level cooperation.
We introduce a novel method based on a 3D Convolutional Neural Network (CNN) to detect starting motions on image sequences by learning spatio-temporal features. The CNN is complemented by a smart device based starting movement detection originating from smart devices carried by the cyclist. Both model outputs are combined in a stacking ensemble approach using an extreme gradient boosting classifier resulting in a fast and yet robust cooperative starting movement detector. We evaluate our cooperative approach on real-world data originating from experiments with $49$ test subjects consisting of $84$ starting motions. 

\end{abstract}



%
\IEEEpeerreviewmaketitle



\section{\large Introduction}
\label{sec_introduction}
\subsection{Motivation}

In our work, we envision a future scenario~\cite{BRZ+17} where traffic participants with different degrees of automation share a road with vulnerable road users (VRUs) such as pedestrians and cyclists. To ensure safe interactions between automated vehicles and VRUs, it is crucial to predict the intentions of the VRUs as early as possible, so that the vehicle is able to choose an appropriate reaction and avoid dangerous situations. However, anticipating the behavior of VRUs is challenging, as they are able to suddenly change their directions. Additionally, they can be overlooked easily due to occlusions. To guarantee safety, automated road users have to maintain a digital map of their surroundings using sensor systems such as cameras, LiDAR, and RADAR. Due to sensor limitations and occlusions, it is also necessary to communicate between different traffic participants to cooperatively update the environment model maintained by each agent. In addition, the model may be complemented by infrastructural sensor systems, such as stationary cameras and laser scanners or by VRUs equipped with mobile devices, such as smartphones or smart watches. Approaches using body worn sensors to locate VRUs often lack positional accuracy, but  allow for an early intention detection. Infrastructures equipped with sensors can help to resolve occlusions. By combining this information we are able to gain a more reliable model of the current traffic situation. The collective knowledge can be utilized in different ways, e.g., to improve the localization of other traffic participants. 

\subsection{Main Contributions and Outline of this Paper}
Our main contribution is a cooperative approach for early and robust starting movement detection of cyclists. We introduce a new method based on a 3D Convolutional Neural Network to detect starting motions on image sequences by learning spatio-temporal features.
Moreover, we present an extended method for starting movement detection based on human activity data, originating from smart devices. Starting movement detections from a camera and smart device based detectors are combined using a novel boosted stacking ensemble approach, implementing decision- and feature-level cooperation.

The remainder of this article is structured as follows: In Sec.~\ref{sec_relatedwork}, the related work in 
the field of cooperative intention detection methods, convolutional neural networks, and cooperative methods is presented. Sec.~\ref{sec_method_overview} describes the overall cooperative approach, the 3D CNN and the smart device based starting movement detection method. In Sec.~\ref{sec_evaluation}, the metrics and data used for evaluation are presented. The experimental results are discussed in Sec.~\ref{sec_ResultsOutline}. Finally, in Sec.~\ref{sec_conclusion} the main conclusions and open challenges for future work are presented.

\section{Related Work}
\label{sec_relatedwork}

\subsection{Intention Detection of Vulnerable Road Users}

The detection of VRUs has been an active field of research over the past decades. But with increasing automation in vehicles it becomes more and more important to not only know the current position of the VRU, but to anticipate their behavior and forecast their future positions. 

We see the motion of a VRU as a sequence of activities or basic movements, such as standing or walking.  Moreover, we see the motion of a VRU as motion of certain body points (e.g., center of gravity, joints, or head) in the 3D space. Our second aim is to forecast trajectories of such points. Both, basic movement detection and trajectory forecasting, are part of what we refer to as intention detection. In this article 
we focus on detecting the starting movement. 

Regarding pedestrians' intentions, research has become more active over the last years. 

In \cite{Koehler.2013}, K\"ohler et al. used Histograms of Oriented Gradients features on motion history images of pedestrians in combination with a Support Vector Machine to detect their intentions to cross the road. The image sequences were generated with a stationary camera. They were able to detect the starting intention within the first step of the pedestrian with an accuracy of 99\%. In \cite{Koehler.2015}, they adapted their method to moving vehicles and different movement types.

Quintero et al. \cite{quintero.2015} presented a method for pedestrian intention and pose prediction based on the pedestrian's 3D joint positions in combination with Balanced Gaussian Process Dynamical Models and na\"ive-Bayes classifiers. They were able to provide path predictions at a time horizon of 1 second with mean errors of 24.4 cm for walking, 26.7 cm for stopping, and 37.4 cm for starting pedestrians.

When it comes to cyclist intention detection, there is still few research in the field. Pool et al. \cite{Pool.2017} presented a method for cyclist path prediction using the local road topology as additional information. They tested their algorithm on real world data recorded from a moving vehicle. By mixing different motion models for canonical directions, they were able to improve the prediction for sharp turns by 20\% on average.

Concerning cooperative VRU intention detection approaches including smart devices, Thielen~et~al. presented a prototype system incorporating a vehicle with the ability of C2X communication and a cyclist with a WiFi enabled smartphone in~\cite{Thielen2012}. The authors were able to successfully test a prototype application that warns a vehicle driver against a likely collision with a crossing cyclist within the next $5\,$s. The authors in~\cite{Engel2013} proposed a prototype system using a smart device for Car2Pedestrian communication and pedestrian tracking. For this purpose, they transmitted the type of movement to an approaching car. This allows for warning the pedestrian and the driver. In~\cite{MLT+17}, the authors presented an approach 
to pedestrian path prediction using artificial neural networks. These three approaches have in common that 
their predicted VRU trajectory is based on smart device information. However, there is no cooperation on the level of detected intentions between different traffic participants.
In~\cite{SSM17}, the authors proposed a concept and listing of requirements for cooperative intelligent transportation systems including smart devices.

 In our own preliminary work \cite{BZD+17}, we introduced a cooperative two stage method to detect starting motions of cyclists and forecast their future positions. The detection of starting motions was done using smart devices and infrastructure based sensors. The detected movement primitives were used in an adaptive gating function that weights two trained trajectory forecasting models. Using this cooperative method, we achieved a more robust detection of starting motions and reduced the forecasting error of starting trajectories.
 
 In this article, we extend our previous method by a 3D Convolutional Neural Network (CNN) incorporating spatio-temporal features and a stacking ensemble approach realizing feature- and decision-level cooperation for cyclist starting intention detection bringing us one step closer toward the envisioned future traffic scenario~\cite{BRZ+17}.

\subsection{Convolutional Neural Networks}
\label{subsec_relatedwork_cnns}
CNNs were first used for visual pattern recognition tasks in the early 90s and have drawn attention to themselves when Krizhevsky et al. \cite{Krizhevsky.2012} used a deep CNN called AlexNet to win the ImageNet Large-Scale Visual Recogintion Challenge (ILSVRC) \cite{imagenet} in 2012. Now they are widely used for image classification and image based object detection. The AlexNet consists of a simple layout where convolutional layers, together with max pooling layers to reduce the dimensionality, and dropout layers to prevent overfitting are stacked on top of each other.

In 2015, Szegedy et al. \cite{Szegedy.2015} introduced the use of so called inception modules where in one layer multiple filter operations are performed in parallel and the output is concatenated at the end. They created a network called GoogLeNet by stacking inceptions modules on top of each other, resulting in a deep architecture with 12 times fewer parameters than AlexNet. GoogLeNet won the ILSVRC 2014.

The ILSVRC 2015 was won by He et al. \cite{He.2015} using a residual network architecture (ResNet). ResNet consists of residual blocks, where the output of convolutional layers is added to their input allowing much deeper network architectures. In this article, we use a ResNet architecture to classify starting movements. The network is described in detail in Sec. \ref{subsec:movement_primitive_detection_cnn}. 

Ji et al. \cite{Ji.2012} developed a CNN based model for human action recognition in video sequences. When it comes to action recognition, the information contained in a single image is often not enough to detect a certain action.
Therefore, they performed 3D convolutions to extract spatial and temporal features from sequences. The action recognition was performed on surveillance videos and the authors compared their method to a classification using a 2D CNN and four other methods using spatial pyramid matching, where the 3D CNN outperformed all other methods.

A different approach using Long-term Recurrent Convolutional Networks, presented by Donahue et al. \cite{donahue.2015}, uses multiple 2D CNNs on single input frames of a video sequence to extract visual features for every time step, which are then passed to a Long Short-Term Memory (LSTM) encoder. An LSTM decoder was used to generate a sentence in natural language describing the video.

Malchanov et al. \cite{Molchanov.2015} introduced a method to classify hand gestures in video sequences using a 3D CNN. By creating two subnetworks, one low and one high resolution network, and fusing the outputs at the end, they were able to achieve a higher accuracy compared to the unfused outputs of the subnetworks. 

\subsection{Cooperative Methods}
\label{subsec:coop_methods}

In this section, the related work on cooperative methods is reviewed.

According to~\cite{KKK+13}, multisensor data fusion is a process which enables the combination of information from different sources in order to form a unified picture. Cooperation refers to the process of acting together for a common benefit and in contrast to bare data fusion also captures the interactions between the different participants which are inherent in our envisioned future traffic scenario~\cite{BRZ+17}. For this reason, we use cooperation as an umbrella term including fusion as an integral part.

Speaking about cooperative methods we have to specify at which level we aim to cooperate: (1) at the data- or feature-level, (2) at the level of 
models and (3) at the level of model outputs or decisions. Our approach focuses on the first and third level. 

In the machine learning research community ensemble techniques~\cite{Zho12} are a widely applied to combine different model outputs in order to improve the overall prediction quality. 
Well known ensemble techniques are stacking, boosting, bagging or Bayesian model averaging~\cite{Bishop2006PRM}. An extension to the latter is Bayesian classifier combination~\cite{KG12}. 
Recently, mixture of experts models, which comprises a gating model weighting the outputs of different submodels, gained a lot of attention, as they determine state-of-the-art performance in language modeling~\cite{SMM+17} and multi-source machine translation~\cite{GC16}. These approaches are based on deep neural networks, hence, they require many training samples.

Besides machine learning based techniques there are also Bayesian filtering~\cite{MRS+14} and sequential estimation techniques~\cite{MD04}, which are mostly used to conduct feature-level fusion. Additionally, Dempster-Shafer theory~\cite{AHS08} is widely used in research and industry for cooperative systems fusing various sources of information. A review over techniques for data fusion in intelligent transportation systems can be found in~\cite{FK16}.

\section{\large Method}
\label{sec_method_overview}

Our approach aims to detect transitions between waiting and starting as early as possible because an early knowledge about these transitions can support the trajectory forecast~\cite{BRZ+17}. 
The detected movements can then be used by an automated vehicle, e.g., for trajectory planning. 
In this work, we focus on detecting the transition between waiting and starting. 
The detection mechanism has to recognize these transitions within a 
few milliseconds after the first starting movement while retaining robustness, i.e., avoiding false positive starting detections. For a more robust and yet fast starting intention detection we propose a cooperative approach. In our method the output of several classifiers with different strengths and weaknesses are combined in order to receive a reliable and fast detector for starting movements of cyclists. The cooperating traffic participants, which we consider in the following as agents, can be cars, infrastructure and VRUs themselves. Here, we restrict ourselves to 
a research intersection~\cite{GoldhammerIntersection.2012} equipped with cameras and smart devices carried by VRUs.

\begin{figure}
	\centering
	\includegraphics[width=0.52\textwidth, clip, trim=110 0 110 20]{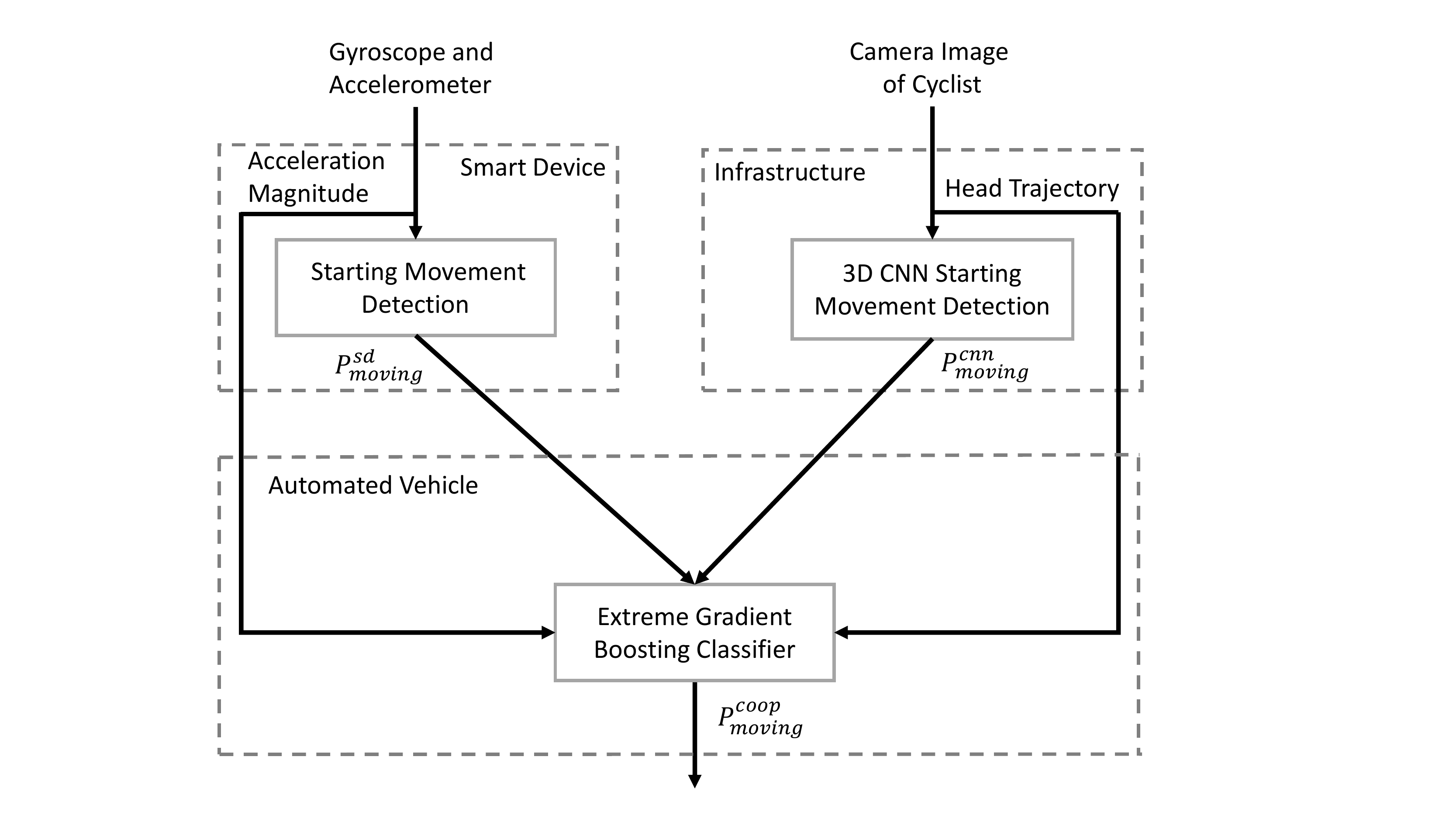}
	\caption{Exemplary cooperative movement detection process incorporating the 3D CNN and smart device based detectors and a cooperative detector running on an automated vehicle.}
	\label{fig:CooperativeIntentionDetection}
	\vskip -4mm
\end{figure}

\subsection{Overall Approach} 
In this section, we describe our overall approach to cooperative starting movement detection.
An exemplary scheme of our cooperative detection process including smart devices, infrastructure and automated vehicles is depicted in Fig.~\ref{fig:CooperativeIntentionDetection}. 
In order to use all available sources of information and to get an encompassing model of the environment, we perform cooperation on two different levels, i.e, the feature- and decision-level. The smart devices and the infrastructure cameras are used to cooperatively detect the cyclist's movement type, i.e., \textit{waiting} or \textit{moving}.

First, an individual detection step is performed by the different agents. 
The detector for the camera-based recognition is realized using a novel 3D CNN based approach. The smart-device based detector is based on human activity recognition techniques using accelerometers and gyroscope measurements. Then subsequently, the detections of each agent are sent to another agent, e.g., to the automated vehicle, where a stacking ensemble approach using extreme gradient boosting is used to combine both detection results. In Fig.~\ref{fig:CooperativeIntentionDetection}, $P_{moving}^{sd}$, $P_{moving}^{cnn}$ and  $P_{moving}^{coop}$ denote the probabilistic outputs for the \textit{moving} class of the detectors of the smart device, 3D CNN and ensemble classifier, respectively.
The stacking ensemble approach realizes cooperation on the level of detected starting movements as well as feature-level cooperation. The latter is the concatenation of features originating from the smart device and the infrastructure camera system. These features, the results of the 3D CNN and smart device based classifiers are used as input of an extreme gradient boosting  classifier, implementing the cooperative starting movement detection.

Moreover, as input for our approach we assume that a mechanism for camera based cyclist recognition, tracking of 2D bounding boxes, 
and 3D head trajectories, e.g., a wide-angle stereo system, is given. For the communication we assume that it is realized by means of an ad hoc network. Our approach assumes idealized communication without any considerable communication delays and synchronized devices using GPS timestamps. The cooperation mechanism avoids sending large amounts of raw sensory data, only the accelerometer's magnitude, the cyclist's head position and the current detector output have to be transmitted. This allows to cope with a limited bandwidth concerning the communication medium.

First, the general modeling concerning the different classes used by the subsequent classifiers for starting movement detection is described. In the following, the movement detection using a sequence of camera images and a residual 3D CNN is described in Sec. \ref{subsec:movement_primitive_detection_cnn}. In Sec. \ref{subsec:movement_primitive_detection_sd}, we present the starting intention detection based on the sensors of the smart device.
Finally in Sec. \ref{sec:cooperative_mp_pred}, the cooperation mechanism realized by means of the stacking ensemble model is described.

\subsection{Detection of Starting Movements}
\label{subsec:3dcnn_starting_movement_classifcation}

The starting movement detection is modeled as a three class problem, i.e.,
\textit{waiting}, \textit{starting} and \textit{moving}. 
The \textit{starting} class is an auxiliary class, which 
allows to integrate early movement indicators, such as head movements, into the training process. Moreover, it simplifies the 
modeling of the transition between the \textit{waiting} and \textit{moving}
class. $P_{waiting}$, $P_{starting}$, and $P_{moving}$ denote the probabilities assigned by the classifiers to the different classes. The labels of the output data were created manually and are defined as follows: A sequence is labeled as \textit{waiting} if neither the wheel of the bicycle is moving, nor the cyclist is performing a movement that leads to a starting motion. Every frame between the first visible movement of the cyclist which leads to a start and the first movement of the wheel of the bicycle is labeled as \textit{starting}. Finally, every frame after the first movement of the bicycle wheel is labeled as \textit{moving}.

\begin{figure}
	\centering
	\includegraphics[width=0.30\textwidth]{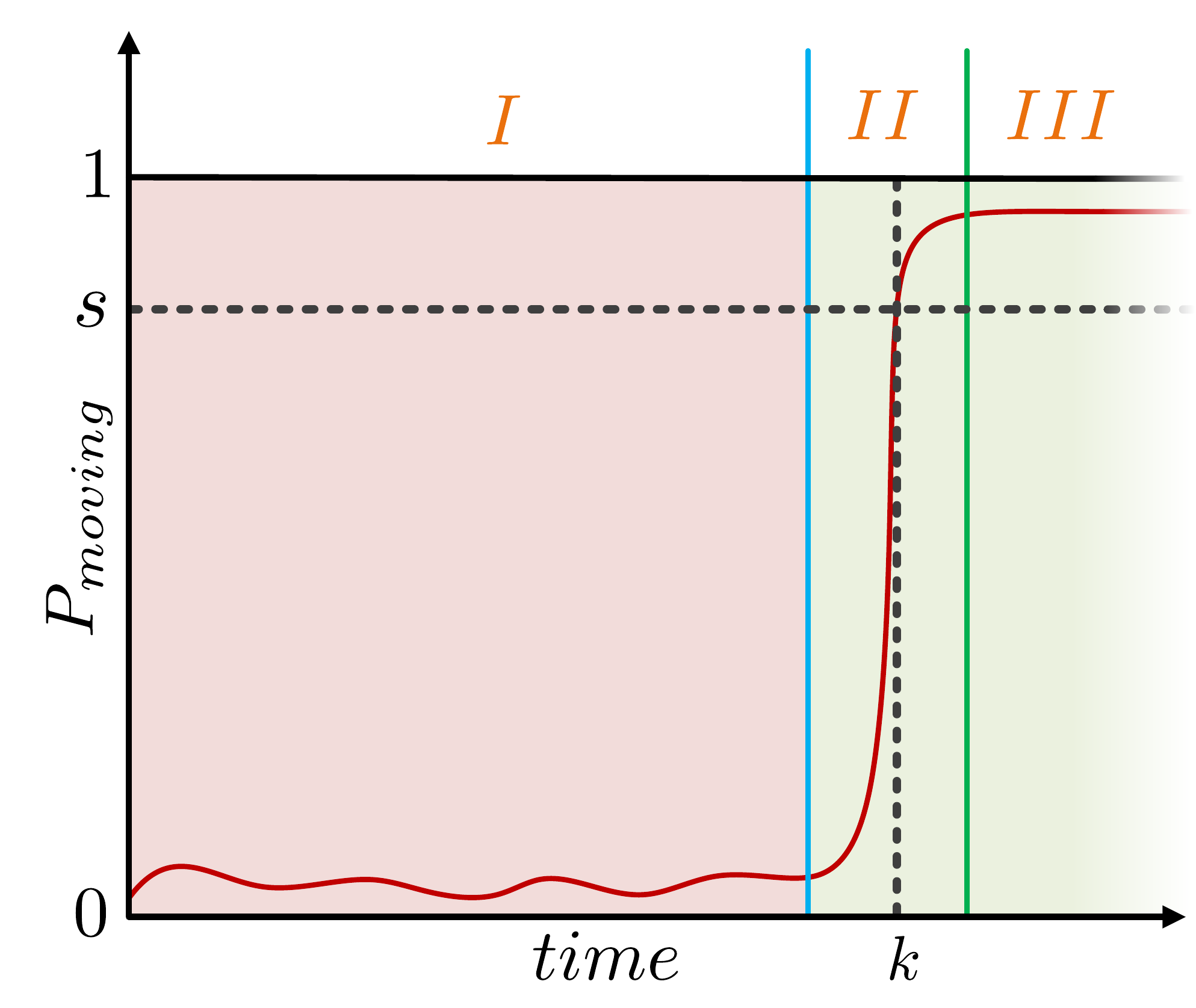}
	\vskip 2mm
	\caption{Exemplary classification output of one scene, with the moving probability $P_{moving}$ (red), the labeled \textit{starting} time (blue), and the labeled \textit{moving} time (green).}
	\vskip -5mm
	\label{fig:ex_net_out}
\end{figure}

Fig. \ref{fig:ex_net_out} shows an exemplary output of the classifier, where the red line represents $P_{moving}$, the blue line is the labeled starting frame and the green line is the labeled moving frame. The output of the classifier should be a zero \textit{moving} probability during phase I (see Fig. \ref{fig:ex_net_out}), the probability should increase in phase II, and should reach a probability of one in phase III. However, sometimes there is no phase II, since there is no visible movement of the cyclist until the wheel of bicycle starts moving.

\subsection{Movement Primitive Detection Using a 3D CNN}
\label{subsec:movement_primitive_detection_cnn}

In this section, we present the use of a 3D CNN with the ResNet architecture to classify starting movements of cyclists. The classification is done frame by frame using the last ten images of a stationary camera.

\subsubsection{Convolutional Neural Network}
\label{subsubsec:cnn}

A CNN architecture mainly consists of convolutional layers which perform convolutions to extract local features from the input and generate feature maps $v_{xy}$, pooling layers to reduce the dimensionality of the feature maps, and a fully connected layer (FCN) at the output.

\small
\begin{equation}
v_{xy} = b_{xy}+\sum_{m=-M}^{M}\sum_{n=-N}^{N}I(x-m, y-n)\cdot K(m+M, n+N)
\label{eq:2dconv}
\end{equation}
\normalsize

Eq.~\ref{eq:2dconv} describes a 2D convolution at the image position $(x, y)$ with the input image $I$ and the filter kernel $K$. The size of the filter kernel in $x$- and $y$-direction is described by $(2M+1, 2N+1)$ for kernels of odd size. After a bias $b_{xy}$ is added to the result of the convolution an activation function, in this case a rectified linear unit (ReLU), is applied. The ReLU outputs $0$ for values smaller than 0, or the value itself for values greater than 0 (Eq.~\ref{eq:relu}):
	
\begin{equation}
f(x)=max(x,0)
\label{eq:relu}
\end{equation}
	
\subsubsection{3D Convolutional Neural Network}
\label{subsubsec:3dcnn}

\begin{figure}
	\centering
	\includegraphics[width=0.49\textwidth]{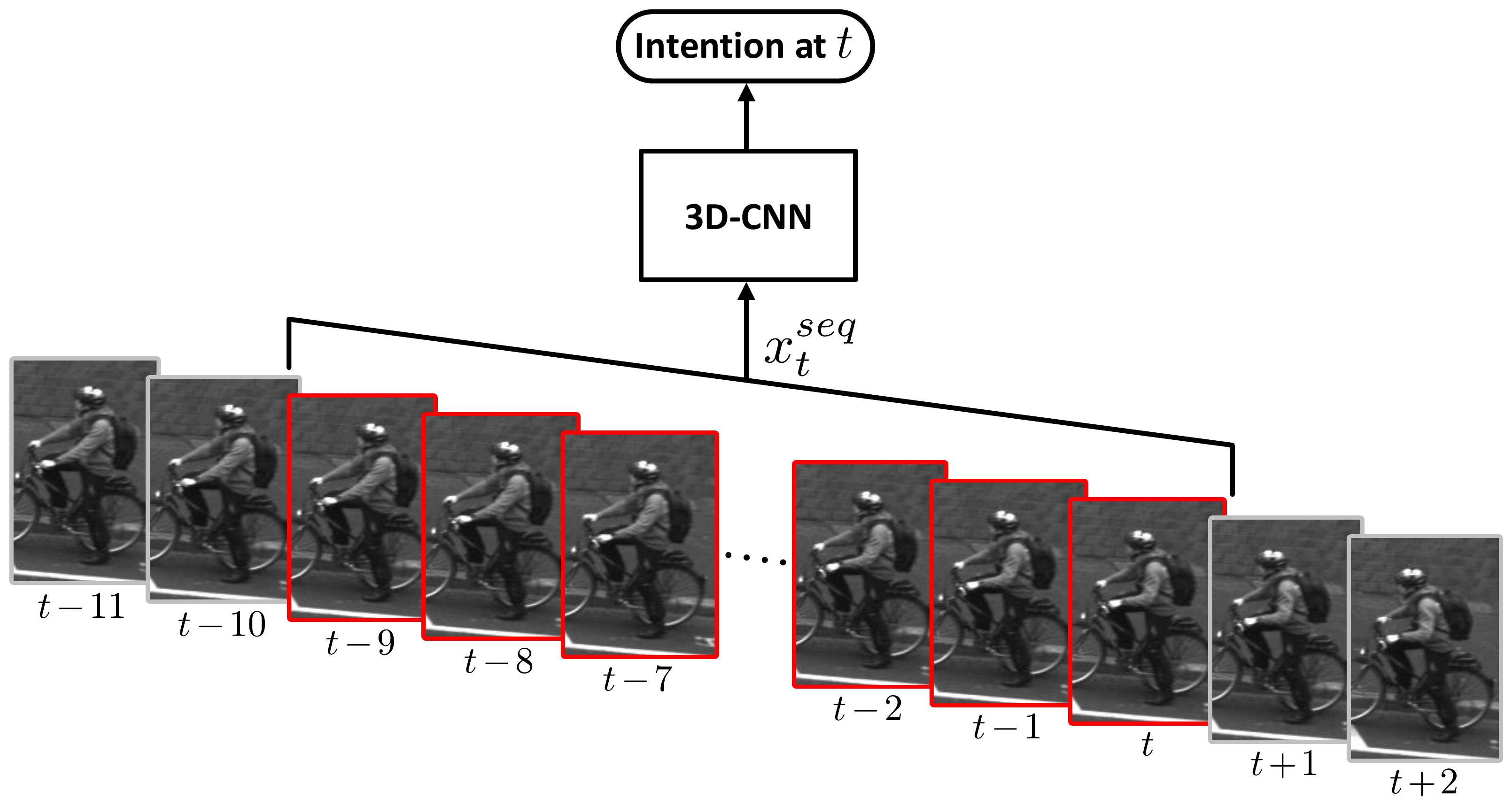}
	\vskip 2mm
	\caption{Example input sequence of 3D CNN containing the last 
	ten images for time $t$.}
	\label{fig:3dcnn_input}
	\vskip -3mm
\end{figure}

Since starting movements of cyclists cannot be captured by a single image, 
we added the temporal dimension to the input of the CNN by stacking the last 10 images ($I_{t-9}$ to $I_{t}$) and adding a third dimension to the filter kernel. Fig. \ref{fig:3dcnn_input} shows an example of the 3D CNN input $x^{seq}_t$. The functional behavior of the 3D convolution is shown in Eq.~\ref{eq:3dconv}, where $t$ is the temporal dimension of the image sequence and $2T + 1$ describes the size of the filter kernel in the temporal dimension. By applying the 3D kernel to an image sequence, the extracted features contain information about movements at a certain position.

{\small
\begin{multline}
v_{xyt} = b_{xyt}+ \\ \sum_{m=-M}^{M}\sum_{n=-N}^{N}\sum_{j=-T}^{T}I(x-m, y-n, t-j)\cdot K(m+M, n+N, j+T)
\label{eq:3dconv}
\end{multline}}
\normalsize

\subsubsection{Residual Architecture}
\label{subsubsec:res}

\begin{figure}
	\centering
	\includegraphics[width=0.49\textwidth]{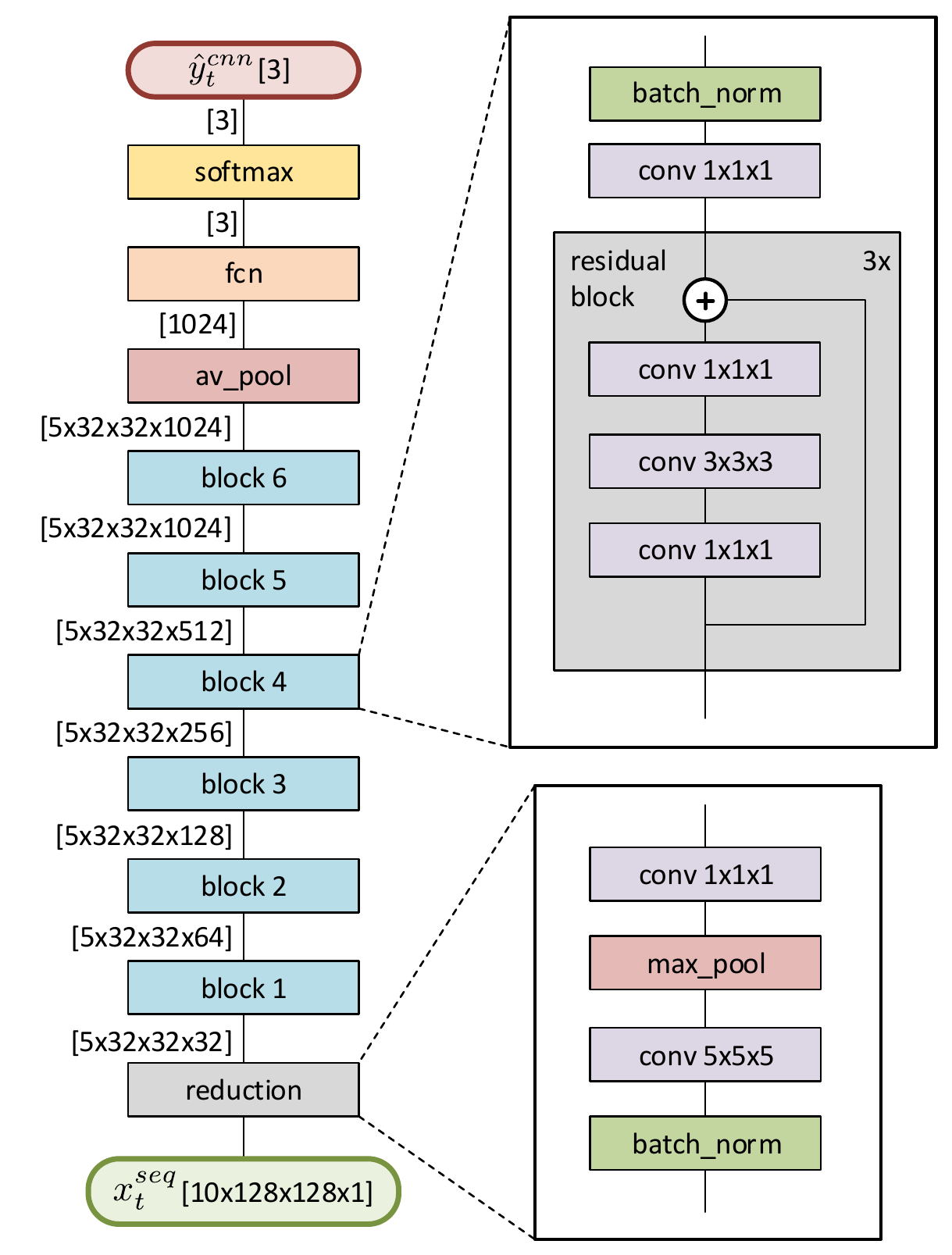}
	\vskip 2mm
	\caption{Architecture of the residual 3D CNN. A residual block is depicted in the upper right and the reduction layer in the lower right.}
	\label{fig:net_arch_1}
	\vskip -3.5mm
\end{figure}

The ResNet architecture was introduced by He et al. in 2015 \cite{He.2015}. They propose a network architecture which is easier to train and produces a higher accuracy than conventional CNNs. The authors identified a degradation problem, where adding more layers to a deep model leads to a higher training error. The problem was addressed by adding residual building blocks, where the original input $x$ bypasses a series of convolution layers $F(x)$ and is added to the output resulting in $F(x)+x$. Thus, instead of directly modeling a target function they force the network to model the residual function. By stacking residual blocks, they were able to train a network with 152 layers resulting in substantially better classification results compared to shallower networks. A residual block is depicted in Fig. \ref{fig:net_arch_1}, on the right. Each block has an internal bottleneck structure to reduce the amount of computation. The first $1\times 1\times 1$ convolution represents the input in a lower dimension, followed by an $N\times M\times T$ convolution.

Our network architecture is described in Fig. \ref{fig:net_arch_1} and is based on the network from \cite{He.2015}. To reduce the number of input features, after a batch normalization layer, an initial convolution followed by a max pooling layer is applied. Before the feature maps are passed to the residual blocks, a $1\times 1\times 1$ convolution is carried out to generate 32 feature maps. A residual block consists of 3 sequential repetitions of residual layers with bottleneck architecture. After 6 blocks, resulting in 1024 feature maps, an average pooling is applied, where the result is passed to a fully connected layer followed by a softmax layer to generate probabilities. In order to speed up the training process, batch normalization layers were added at the network input and after every residual block.

The output vector $\hat{y}_t^{cnn}$ of the overall network contains the probabilities of 3 classes: $P_{waiting}$, $P_{starting}$, and $P_{moving}$. 
Our network is trained on the input sequences $x^{seq}_t$ of each individual time step and outputs a feature vector $\hat{y}_t^{cnn}$, which contains the class probabilities.

\subsection{Movement Primitive Detection Using Smart Devices}
\label{subsec:movement_primitive_detection_sd}

\begin{figure*}
	\centering
	\includegraphics[width=1.7\columnwidth, clip, trim=0 60 0 105]{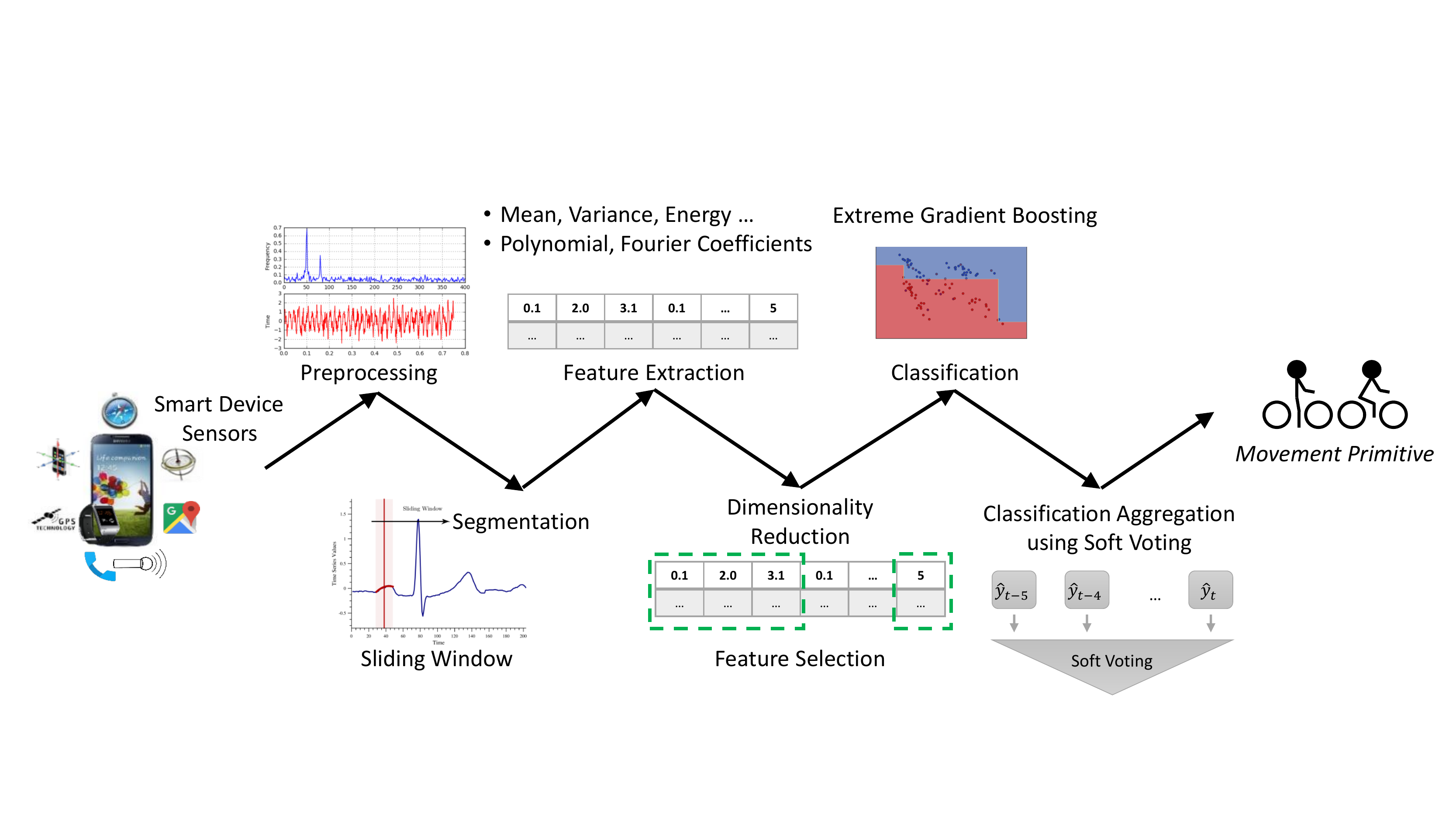}
	\caption{Process for starting movement detection based on smart devices. 
		It consists of the six stages, preprocessing including coordinate transformations, segmentation using a sliding window approach, 
		feature extraction and feature selection, then classification and aggregation via a soft voting scheme.
	}
	\label{fig:recognition_pipeline}
	\vskip -4mm
\end{figure*}

In this section, the smart device based starting movement recognition is described.
The focus of this section is on the early detection of cyclist movement types. 
The detection is realized by means of human activity recognition techniques \cite{Bulling2014THA} based on the gyroscope and accelerometer sensory data, which is nowadays available on nearly every smartphone. A schematic of the performed steps for starting movement detection based on smart devices is depicted in Fig.~\ref{fig:recognition_pipeline}.

\subsubsection{Preprocessing and Feature Extraction}
\label{subsubsec:sd:preprocessing_feature_extraction}
The approach presented here uses features computed from accelerometer and gyroscope sensors sampled 
with a frequency of \SI{100}{\Hz}. The gyroscope allows for detecting rotation movements, such as pedaling, whereas the accelerometer is better suited to detect linear movements, e.g., forward movements. The accelerometer sensor is gravity compensated.
The three accelerometer and gyroscope components ($x$, $y$, and $z$, respectively) are transformed using the estimated gravity vector (e.g., obtained by low-pass filtering of accelerometer data).
The transformed values are then in a coordinate frame which is leveled with the local earth ground plane, i.e., the $z$-axis is pointing towards the sky. This coordinate frame is referred to as local frame. The compass is not considered due to its sensitivity to a precise calibration~\cite{MGF+17} and possible magnetic perturbations. In absence of any compass data, the transformation from this local frame to a global reference frame is not known, i.e., it is not known how the device is oriented with respect to the VRU. By considering the magnitude of the accelerometer and gyroscope values in the local horizontal $x-y$ plane orientation invariance is achieved.
Moreover, the projection of the sensor values on the local vertical $z$-axis, i.e., the gravity axis, are considered.

A sliding window segmentation of window sizes \SI{0.1}{\second} and \SI{0.5}{\second} is performed on each of the transformed signals and features commonly used in human activity recognition~\cite{Bulling2014THA}, such as the mean, variance, and energy, are computed. 
In addition, features based on the orthogonal polynomial approximation up to the $3^{\mathrm{rd}}$ degree are extracted for window lengths of \SI{0.2}{\second} and \SI{0.8}{\second}, as used in~\cite{BZD+17}. Whereas, features computed with the smaller window sizes capture short term dependencies and with larger window sizes capture dependencies on a more coarse timescale.
Additionally, the magnitude of the discrete Fourier transform (DFT) coefficients are also considered as input features, as applied for human walking speed estimation in~\cite{PPC+12}. The coefficients are normalized with respect to the overall energy in the respective window. The window size is set to \SI{0.64}{\second} and coefficients up the $10^{\mathrm{th}}$ order are considered. 
In total, $112$ features are computed.
	
Features extracted from the smart device's integrated GPS are not considered. The reason for this is that GPS is not always available or noisy due to multipath effects. Moreover, the sampling frequency, of the GPS is too low to detect fast changes in the cyclists movement.

\subsubsection{Detection of Starting Movements}
\label{subsubsec:sd:detection_of_starting_movements}
The starting detection is realized by means of a frame-based extreme gradient boosting classification~\cite{CG16}.
The frame-based classification is performed at discrete points with a frequency of \SI{100}{\Hz}. 
To reduce the dimensionality and increase the generalization performance, a sequential forward feature selection procedure is performed. As score, the wrapper approach uses the harmonic mean between the $F_1$-score (i.e., detection 
of starting movements) and the mean time $t_{d}$ required for detecting the starting movement. Whereas the latter is transformed using an exponential transformation $\exp{(-\frac{t_{d}^2}{s})}$ with scale $s = 0.075$ to squash the mean detection time into the unit time interval, i.e., predicting the starting intention exactly at the labeled time results in a score of one.  

The classifier is trained on labeled data consisting of two classes, i.e., \textit{waiting} and \textit{moving}.
The \textit{moving} class as introduced before is merged with the \textit{starting} class for training of the classifier.
The resulting classifier outputs probabilities, which refer to the confidence of the prediction. 
In order to predict proper class probabilities, a probability calibration using a logistic sigmoid is performed.

These probabilities are used in a soft voting ensemble approach~\cite{Zho12},  aggregating the prediction 
of the classifier in a sliding window manner within the last \SI{0.1}{\second}. This smooths the output and reduces false positive detections.

\subsection{Cooperative Movement Primitive Detection}
\label{sec:cooperative_mp_pred}

In this section, we describe the cooperative starting movement detection approach realized by means of a stacking ensemble approach
using another classifier~\cite{Zho12}. 

\begin{figure}[b]
	\centering
	\includegraphics[width=\columnwidth, clip, trim=5 2 5 20]{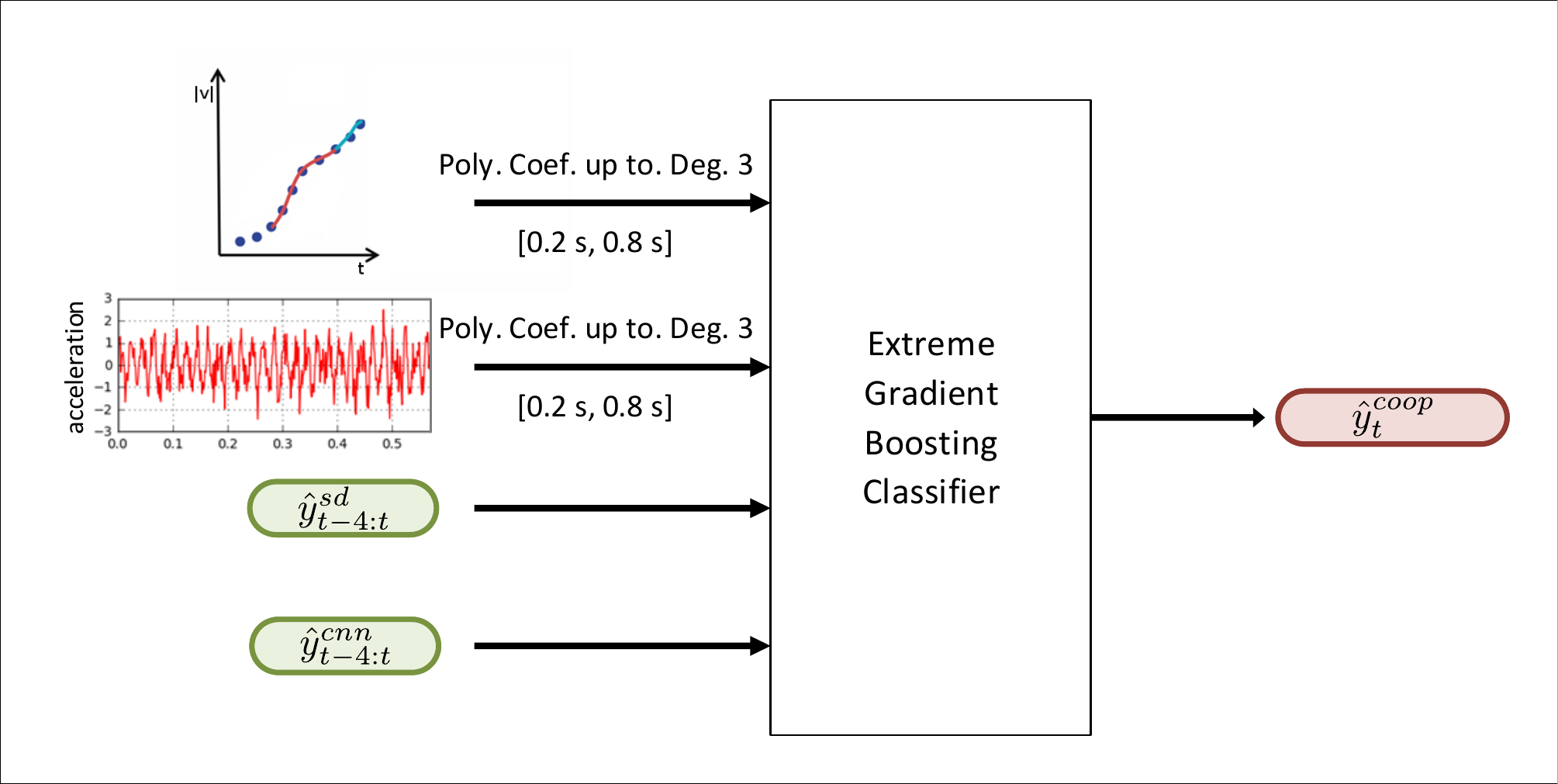}
	\caption{Stacking ensemble approach for cooperative starting movement detection using an extreme gradient boosting classifier.}
	\label{fig:CooperativeIntentionForecast}
\end{figure}

A schematic of this cooperative process is depicted in Fig.~\ref{fig:CooperativeIntentionForecast}.
The classifier combines the detections of the 3D CNN and smart device based detectors. It is realized by an extreme gradient boosting classifier~\cite{CG16}. 

The outputs of the 3D CNN and smart device based classifier ($\hat{y}_t^{\mathrm{cnn}}$ and $\hat{y}^{\mathrm{sd}}_t$, respectively) are used as input of the classifier. In order to increase the performance 
of the classifier, additional features originating from the smart device and the camera are used. Additional smoothing and outlier detection capabilities are added to the classifier 
by consideration of the last three predictions of the 3D CNN and smart device based classifier as inputs.

The features used from the smart device are orthogonal polynomial approximation coefficients~\cite{Fuchs2010} up to the $3^{\mathrm{rd}}$ degree computed on the magnitude of the gravity compensated acceleration values for window lengths of \SI{0.2}{\second} and \SI{0.8}{\second}.
The goal here is not to make extensive use of many features from the smart device, but to give the classifier information to improve the starting movement detection. 
Additionally, using only features computed on a single sensor value reduces the communication overhead. Only the magnitude of the smart device acceleration sensor and the starting movement detection result have to be transmitted.

Moreover, the classifier uses past velocity measurements extracted from the head trajectory of the tracked VRU as input. The information 
can for example be supplied by infrastructure based sensors or vehicles.
As shown in~\cite{Hubert.2017}, these features based on the cyclist's head trajectory are a valuable source of information 
for predicting cyclist intentions. The head trajectory is used to calculate the magnitude of the velocity and subsequently features based on orthogonal polynomial approximations for window lengths of \SI{0.2}{\second} and \SI{0.8}{\second} are extracted.

In order to comply with the sampling rate of the smart devices, the predictions are performed with a frequency of \SI{100}{\Hz}, i.e., the camera measurements are oversampled.
In case of the cyclist being occluded, the cooperative classification stage is bypassed and only the smart device prediction is used.

\section{Data Acquisition and Evaluation}
\label{sec_evaluation}

\begin{figure}
	\centering
	\includegraphics[width=0.49\textwidth]{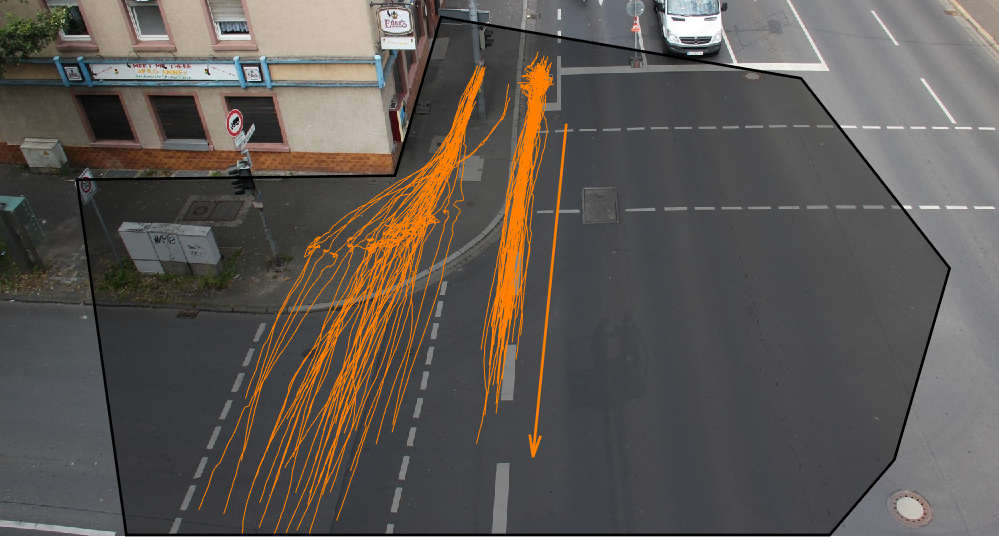}
	\vskip 4mm
	\caption{Overview of the intersection with all starting movements of instructed cyclists. The arrow is pointing into the starting direction.}
	\label{fig:trajectories}
	\vskip -4mm
\end{figure}

\subsection{Data Acquisition}
\label{sec:data_acquisition}

To evaluate our algorithm, we created two datasets of starting cyclists. The first dataset contains 49 female and male test subjects, who were equipped with a smart device. They were instructed to move between certain points at an intersection with public, uninstructed traffic while following the traffic rules. Due to two traffic lights at the intersection, we received 84 starting motions, with a maximum of two starting motions per test subject. The trajectories of the cyclists are shown in Fig. \ref{fig:trajectories}. Besides the sensor data of the smart devices, we recorded the images of two HD cameras, which are installed at the intersection as part of a wide angle stereo camera system. The smart device measurements were used to train, validate, and test the classifier based on data from smart devices. In order to increase the number of validation sequences for the stacking ensemble, the 3D CNN is only trained on uninstructed cyclists and the instructed are only used for evaluation using a two-fold cross-validation.

The second dataset was created to enhance our data and consists of image data  of 305 uninstructed cyclists, recorded at the same intersection. The second set was split into training, validation and test sets for the 3D CNN.

\subsection{Evaluation}
\label{sec_method_evaluation}

We performed the evaluation of our cooperative approach using  infrastructure and smart device based sensors. The evaluation, i.e., 
cooperation, is performed offline.

To assess the quality of our classifiers, we created a scene-wise evaluation, where one scene starts after the cyclist stopped and ends when the cyclist is leaving the field of view. To classify a starting movement, we look at the predicted moving probability. If a certain threshold is reached (\textit{s} in Fig. \ref{fig:ex_net_out}), the starting movement is classified at the time step where the threshold was reached (\textit{k} in Fig. \ref{fig:ex_net_out}). If starting was detected during phase I, the scene is rated as false positive. If the detection is in phase II or  III, it is rated as true positive. If the threshold was never reached, it is rated as false negative. Since every waiting phase ends in a starting phase, we do not consider true negatives.
The overall quality of the classifiers is rated by $F_1$-score and precision.

To assess the detection time of the classifier, we calculate the mean time difference $\overline{\delta_t}$ between the detection time $t_d$ and the start time of phase III $t_{III}$ of all true positives over all $L$  sequences (Eq. \ref{eq:mean_detection_time}), where smaller values indicate a faster classification.

\begin{equation}
\overline{\delta_t}=\frac{1}{L}\sum_{i=1}^{L}(t_{di}-t_{IIIi})
\label{eq:mean_detection_time}
\end{equation}

\section{Experimental results}
\label{sec_ResultsOutline}

In the following, the described methods are evaluated using a two-fold cross-validation over the set 
of instructed cyclists. First, we evaluate the individual results of each classifier in Sec. \ref{sec:results_basic_moevement}. In Sec. \ref{sec:results_cooperative_prediction}, we assess the cooperative movement primitive detection and compare the results to the individual classifiers.

\subsection{Movement Primitive Detection}
\label{sec:results_basic_moevement}

\begin{figure*} [h]
	\centering
	\vskip -4mm
	\includegraphics[width=1.7\columnwidth]{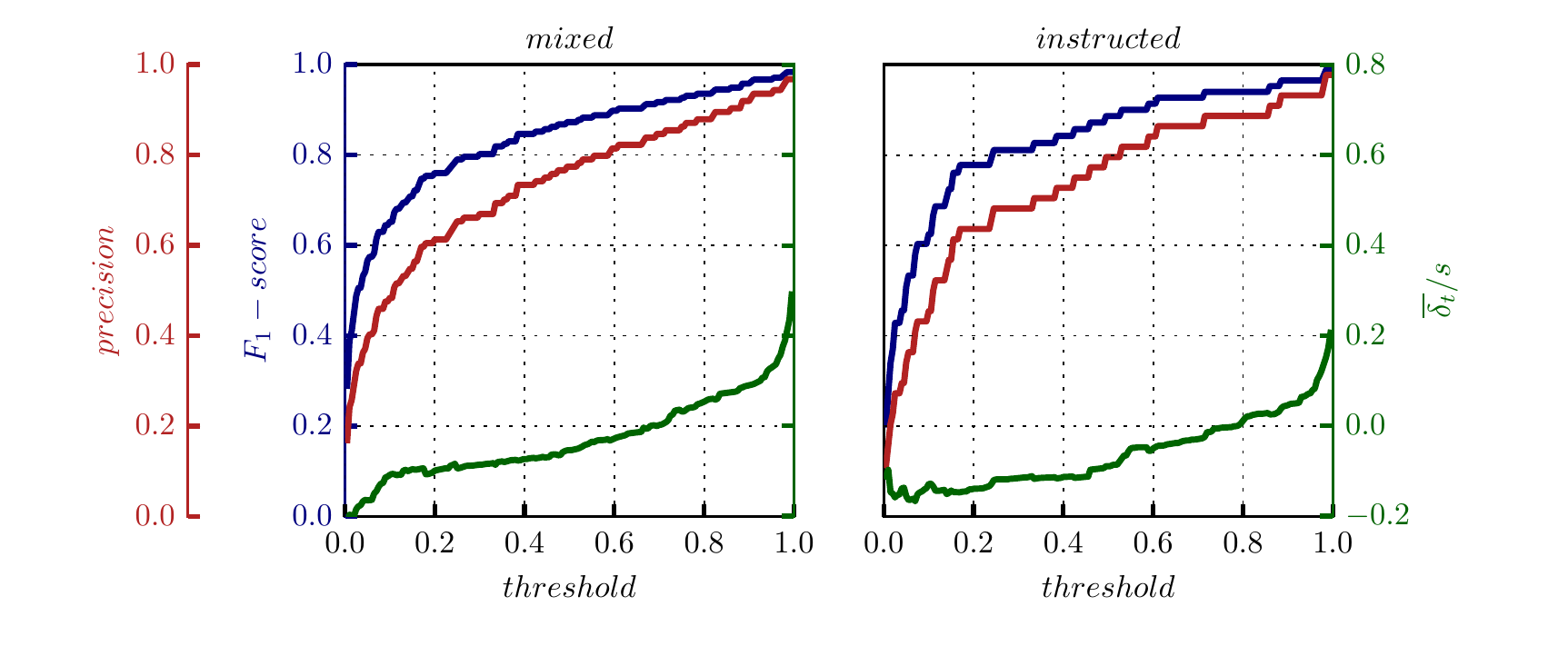}
	\vskip -2mm
	\caption{$F_1$-score (blue), precision (red) and mean detection time (green) over the probability thresholds of all cyclists (left) and instructed cyclists (right).}
	\label{fig:res3dcnn}
	\vskip -2mm
\end{figure*}

\subsubsection{3D CNN}
\label{subsubsec:3dcnn_results}
In this section, we describe the evaluation of the 3D CNN. The dataset we used is divided into a training and a validation set, $s_{train}$ and $s_{val}$. $s_{train}$ consists solely of uninstructed cyclists and makes up 60\% of all measurements. $s_{val}$ consists of all instructed cyclists and additional uninstructed cyclists and was split into two sets $s_{val1}$ and $s_{val2}$ to perform a two-fold cross-validation. $s_{val}$ is split in a way that instructed and uninstructed cyclists are evenly distributed. The classification of the 3D CNN is evaluated on $s_{val}$. To compare the classification results of the 3D CNN to the smart device classifier, we additionally illustrate the results of the instructed cyclists.

The overall results of the classifier are shown in Fig.~\ref{fig:res3dcnn}. To generate the plots, the $F_1$-score, the precision, and $\overline{\delta_t}$ of the classifier are determined (as described in Sec.~\ref{sec_method_evaluation}) for different probability thresholds. The results are plotted over the thresholds from zero to one. The plot on the left contains the results of both folds using all sequences, the right plot contains the results of only the instructed cyclists, respectively. 

Our evaluation shows, that the classifier reaches an $F_1$-score of 90\% at \SI{-0.02}{\second} and 98\% at \SI{0.19}{\second}, on average on the \textit{mixed} data. The evaluation on instructed cyclists produces similar results with an $F_1$-score of 90\% at \SI{-0.07}{\second} and 98\% at \SI{0.14}{\second}. We can see an increase in the $F_1$-score and the precision (blue and red curves) with an increasing threshold due to fewer false positives. The mean detection time $\overline{\delta_t}$ (green) increases with the threshold, because $P_{moving}$ usually increases over time until the maximum probability is reached.

\begin{figure}[h]
	\centering
	\includegraphics[width=0.5\textwidth]{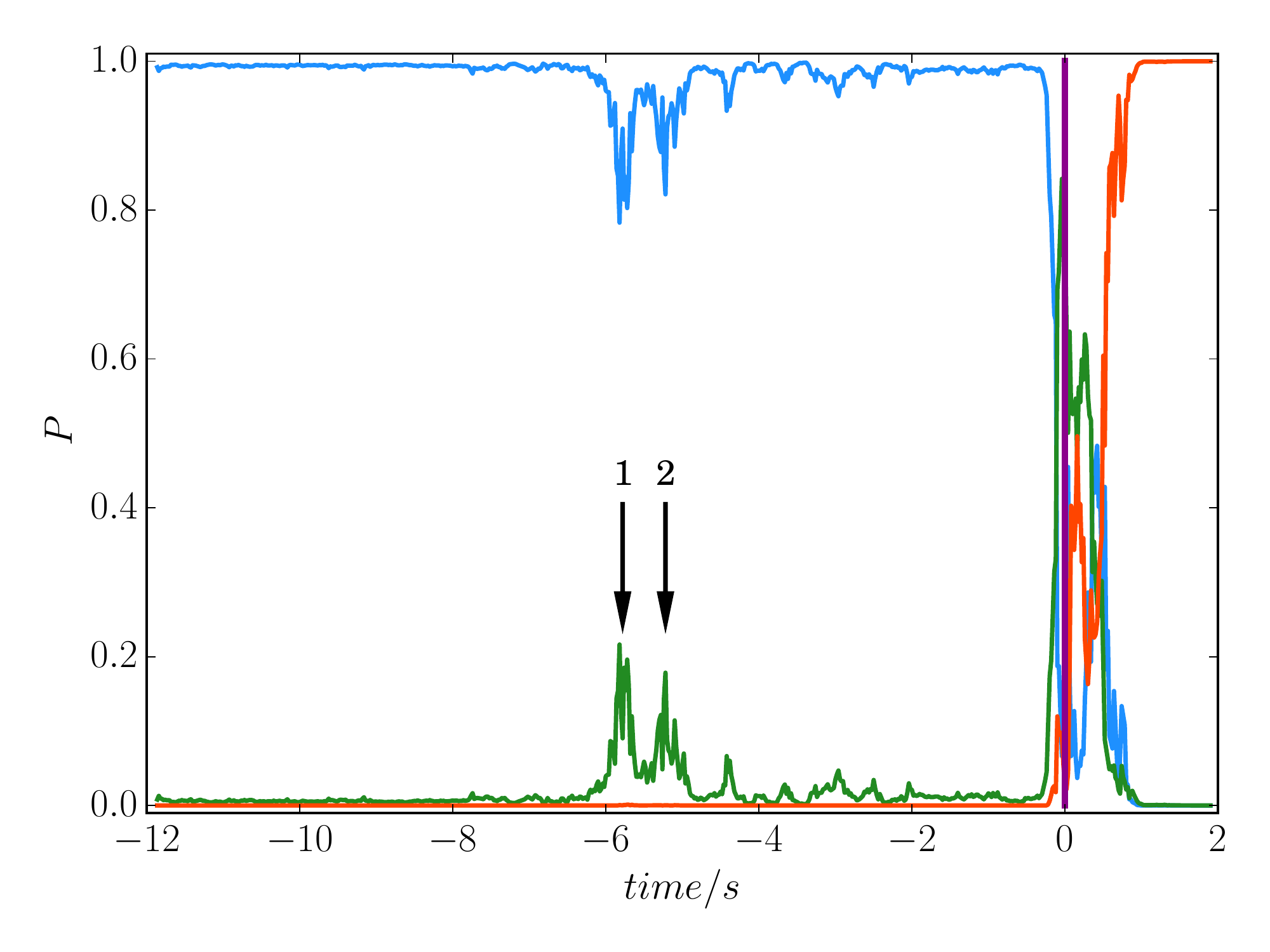}
	\caption{Example classification of one scene with two pedestrians passing in the background of the cyclist. The probabilities $P_{waiting}$ (blue), $P_{starting}$ (green) and $P_{moving}$ (red) are shown. Label $\textit{1}$ references the pedestrian on the left pushing a bicycle. Label $\textit{2}$, the pedestrian in the background on the right.}
	\label{fig:plot_behind}
	\vskip -4mm
\end{figure}

\begin{figure}[b]
	\centering
	\includegraphics[width=0.45\textwidth]{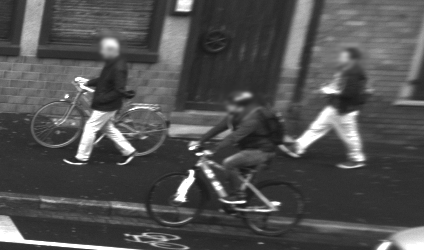}
	\vskip 2mm
	\caption{Two pedestrians passing in the background of the cyclist.}
	\label{fig:grafic_behind}
\end{figure}

Further analyses showed, that the classifier is robust against small movements of the cyclist and even movements in the background of the cyclist, e.g., a pedestrian walking by. Fig.~\ref{fig:grafic_behind} shows a snapshot of a scene. Two pedestrians, one pushing a bicycle, are passing behind the target cyclist. In Fig. \ref{fig:plot_behind}, the corresponding network output is shown. The passing pedestrians cause a slight increase in $P_{starting}$ (green line, peak $1$ and $2$), however, $P_{moving}$ stays unaffected.

\begin{figure}[h]
	\centering
	\includegraphics[width=0.5\textwidth]{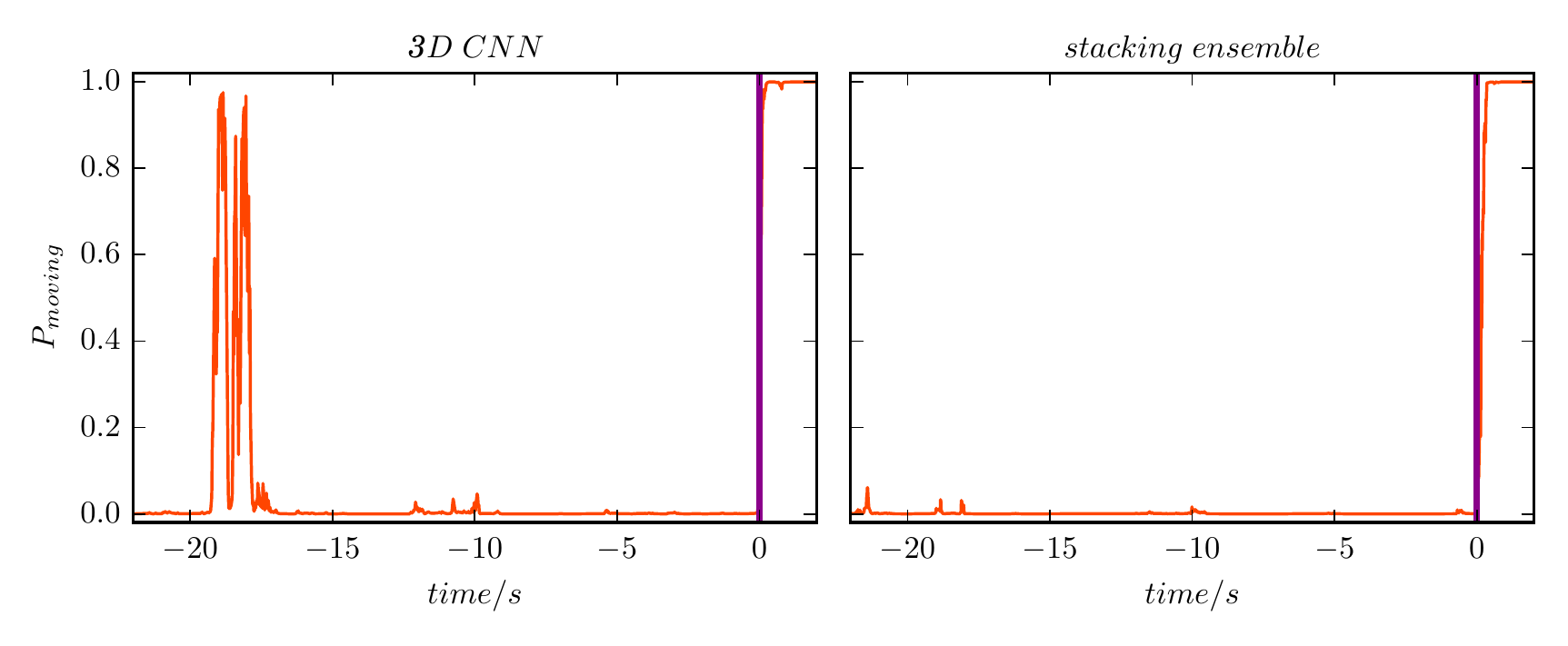}
	\vskip 2mm
	\caption{Comparison between the classification output of the 3D CNN (left) and the stacking ensemble (right). $P_{moving}$ of the 3D CNN is falsified by a pedestrian walking in the background, this is compensated in the stacking ensemble output.}
	\label{fig:failure_plot_behind}
	\vskip -4mm
\end{figure}

\begin{figure}[b]
	\centering
	\includegraphics[width=0.393\textwidth]{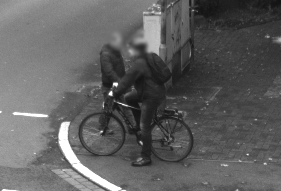}
	\vskip 2mm
	\caption{Pedestrian passing close to the cyclist.}
	\label{fig:grafic_behind2}
\end{figure}

Fig. \ref{fig:failure_plot_behind} (left) shows the network output of a scene with a pedestrian passing very close to the cyclist, which leads to a sharp increase in $P_{moving}$ resulting in a false positive detection. We observed that the network strongly reacts to movements in the area of the bicycle wheels. The passing pedestrian is depicted in Fig.~\ref{fig:grafic_behind2}.

\begin{figure}[h]
	\centering
	\includegraphics[width=0.50\textwidth]{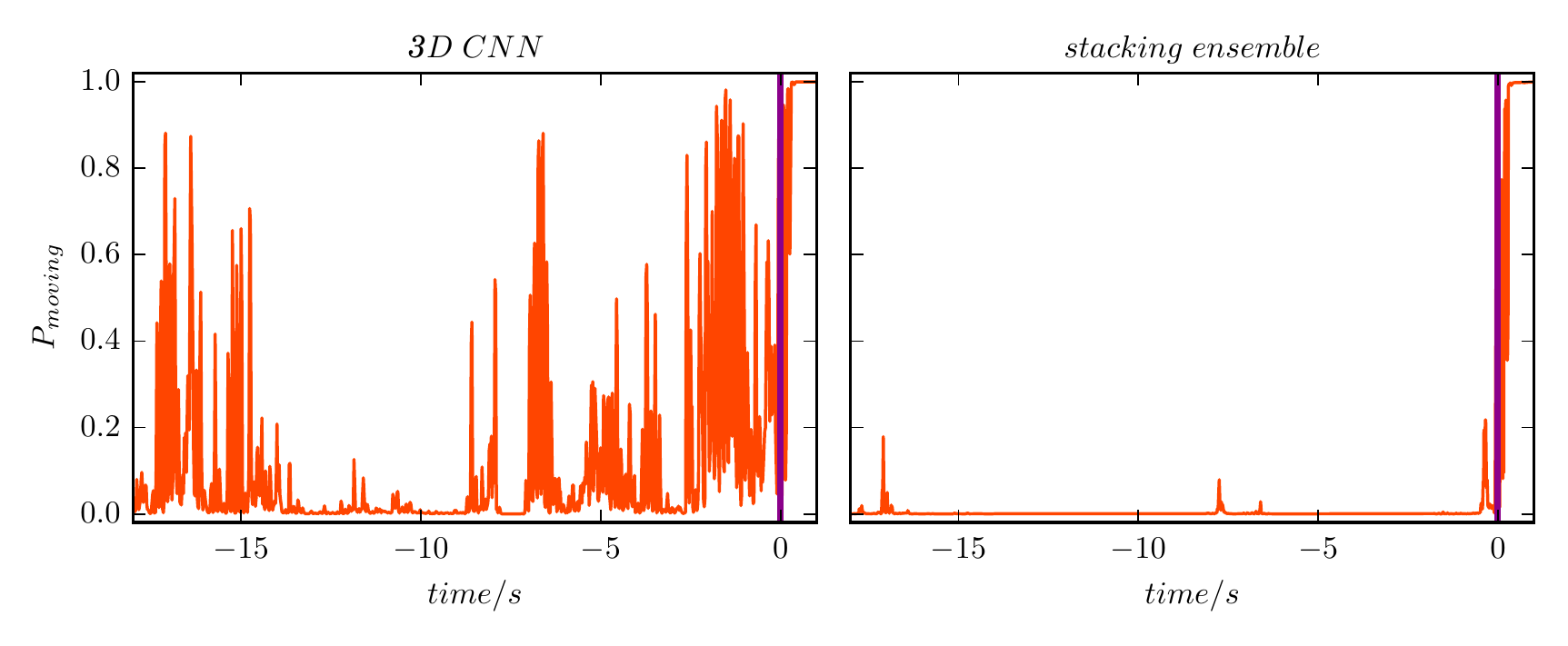}
	\caption{Comparison between the classification output $P_{moving}$ of the 3D CNN (left) and the stacking ensemble (right). Here, the camera shakes due to heavy winds.}
	\label{fig:cmp130}
\end{figure}

Another problem are changing environmental conditions, i.e., occlusion, lighting, or weather 
conditions indirectly affecting the detectors performance. 
In our evaluation data, stormy weather is an example for this changing 
environmental conditions. Here, the otherwise steady camera images start to shake, which leads to a noisy output (Fig.~\ref{fig:cmp130}, left).

Since the smart device sensors are not affected by passing VRUs or changing environmental conditions such as weather situations, we can use the output of the smart device classifier to resolve these issues.




\subsubsection{Smart Devices}
\label{subsubsec:result_smart_devices}
In this section, we describe the evaluation of the smart device based starting movement detector. 
For evaluation of the smart device based approach only the set of instructed cyclist can be considered. 

The overall results of the smart device based detector are shown in Fig.~\ref{fig:final_cmp}. 
Our evaluation shows that the classifier reaches an $F_1$-score of 80\% at a mean detection time of \SI{0.16}{\second}. 
We observe that the precision of the classifier is low compared to the 
the camera based 3D CNN approach.  
The classifier produces many false positives and is sensitive to small motions of the cyclist.
Most of them occur due to movements of cyclists while they are waiting, e.g., when the cyclist is preparing the pedal to start, but is not starting yet. Other false positives 
are due to seesawing movement, i.e., swaying from one leg to the other.
This results in an activity pattern similar to the one observed just before starting.
Many of those false positives can be filtered, i.e., by averaging the last prediction. But this comes at the cost of reduced detection speed.
This shows that for the smart device based detector a high threshold has to be chosen in order to achieve a larger $F_1$-score. Eventually an additional smoothing is necessary.
Despite of the high number of false positive detections, the detection can be a valuable source of information for a cooperative classifier, since it is less affected by changing environmental conditions or passing VRUs.
In case of occlusion the smart devices can be the only source of information left. Hence, the camera and smart device based classifier can complement each other. 

\subsection{Cooperative Movement Primitive Detection}
\label{sec:results_cooperative_prediction}
In this section, we describe the evaluation of the cooperative approach.
As before, the evaluation is performed via a two-fold cross-validation on the set of instructed cyclists.
The smart device and cooperative movement detector have to share the same training set. This is compensated by performing a nested five-fold cross-validation over the VRUs in the respective training fold. 
The smart device based model is fitted on the training data of the nested training folds. The predictions of 
the nested validations folds are concatenated creating the training basis for fitting the stacking ensemble model.

\begin{figure}[h]
	\centering
	\includegraphics[width=0.42\textwidth]{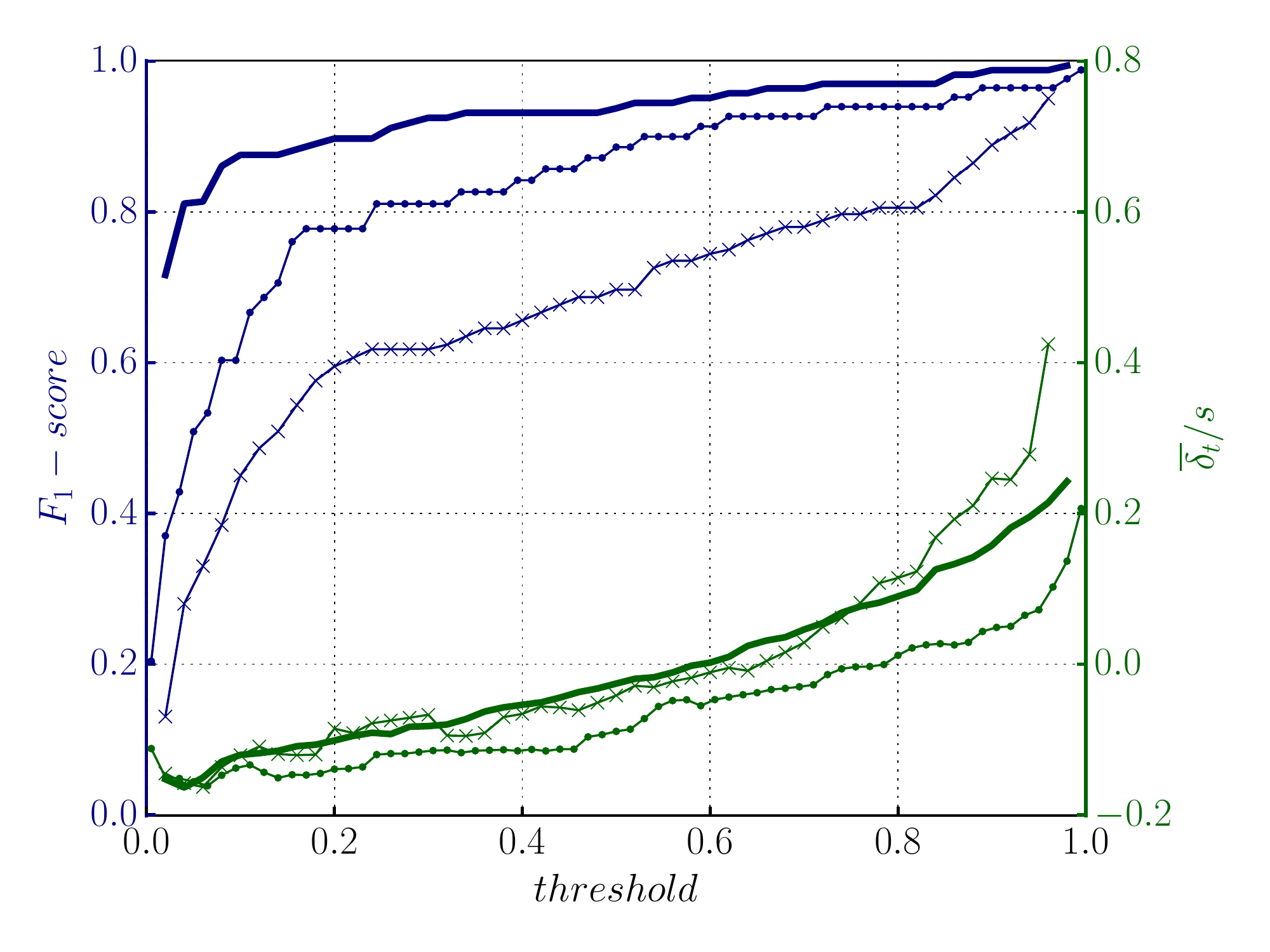}
	\caption{$F_1$-score (blue) and mean detection times (green) over the probability thresholds of the instructed cyclists for the different models, cooperative stacking ensemble (solid), 3D CNN (dotted) and smart device based (crossed).}
	\label{fig:final_cmp}
\end{figure}

The results of the cooperative approach are depicted in Fig.~\ref{fig:final_cmp}.
The classifier reaches an $F_1$-score of 90\% at \SI{-0.09}{\second} and 98\% at \SI{0.13}{\second} 
evaluated on the instructed cyclists. The cooperative stacking ensemble approach outperforms the smart device and 3D CNN based approaches concerning the $F_1$-score. 
The cooperative approach produces an $F_1$-score of $99$\% and a mean detection time of \SI{0.24}{\second} for threshold values greater than $0.98$. 
The cooperative approach combines robustness while retaining a low mean detection time. 

We observe an increased robustness against false positive detections. 
In Fig.~\ref{fig:failure_plot_behind}, this is illustrated for a scene with a 
passing pedestrian behind a waiting cyclist. While there are  
\textit{moving} probability peaks in the solely 3D CNN based approach (Fig. \ref{fig:failure_plot_behind}, left), there 
is no notable peak for the stacking ensemble (Fig.~\ref{fig:failure_plot_behind}, right).

The cooperative approach offers also increased robustness concerning 
environmental conditions, e.g., shaking camera. As depicted in Fig.~\ref{fig:cmp130}, the noise in the \textit{moving} probability predictions is tremendously reduced allowing a clear starting detection. 

\section{\large Conclusions and Future Work}
\label{sec_conclusion}
In this article, we presented an approach to cooperatively
detect starting movements of cyclists. 
The approach consists of a novel 3D CNN based starting movement detector, a smart device based detector, and a stacking ensemble realizing feature- and decision-level cooperation. The latter combines predictions of camera and smart device based detectors.

We showed that the 3D CNN based approach delivers extremely fast and robust predictions, i.e., an $F_1$-score of $90$\% and a mean detection time of \SI{-0.02}{\second}.  
We showed an approach for smart device based starting movement 
detection using human activity recognition techniques. 
The approach delivers promising results, e.g., to cope with occlusion situations.
The cooperative approach combines robustness while retaining a low mean detection time. The stacking ensemble approach combines both estimators, such that they complement each other leading to increased performance, i.e., an $F_1$-score of $99$\% with a mean detection time of \SI{0.24}{\second}. Moreover, we demonstrated the robustness of the single detectors 
and the cooperative approach in sample scenes including passing 
pedestrians and bad environmental conditions, e.g., a shaking camera.

Our future work will focus on extending our presented approach to different traffic scenarios and locations, e.g., cyclist who are turning, or applying the cooperative approach to other VRU types, such as pedestrians. We also intend to include more realistic communication models, e.g., including delays and packet loss.  Moreover, we aim to integrate car based perception. Here, the environmental conditions are even more dynamic, stressing the advantages of cooperative approaches.
We will also investigate on extending the fast and reliable starting movement detection with an advanced trajectory forecasting, e.g., based on deep learning models. Concerning the smart devices we will review different wearing positions and especially explore the effects regarding the movement detection performance. Moreover, we also aim to integrate the smart devices into 
a cooperative tracking mechanism. Additionally, we also aim to work on a smart device based trajectory forecasting. Then we will investigate different types of cooperation, e.g., cooperative trajectory forecasts.


\section{\large Acknowledgment}

This work results from the project DeCoInt$^2$, supported by the German Research Foundation (DFG) within the priority program SPP 1835: ``Kooperativ interagierende Automobile", grant numbers DO 1186/1-1 and SI 674/11-1. Additionally, the work is supported by ``Zentrum Digitalisierung Bayern".



\bibliographystyle{IEEEtran}
%

{\small
\bibliography{IEEEabrv,egbib,sz,mb}
}
\vskip -15mm

\begin{IEEEbiography}[{\includegraphics[width=0.9in,height=1.125in,clip,keepaspectratio]{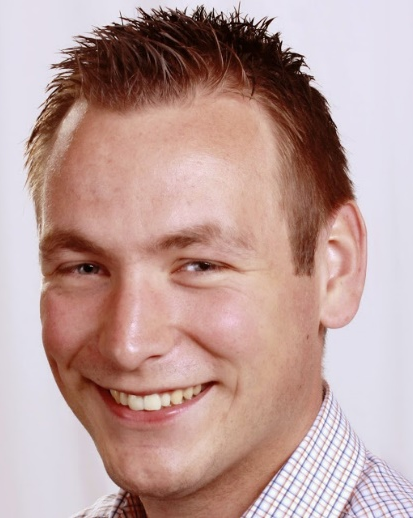}}]
	{Maarten Bieshaar} Maarten Bieshaar received the B.Sc. and the M.Sc. degree in Computer Science from the University of Paderborn, Germany, in 2013 and 2015, respectively. Currently, he is working toward the PhD degree at the University of Kassel, Germany, where he is part of the Intelligent Embedded Systems research group chaired by Bernhard Sick.
\end{IEEEbiography}

\vskip -15mm
\begin{IEEEbiography}[{\includegraphics[width=0.9in,height=1.125in,clip,keepaspectratio]{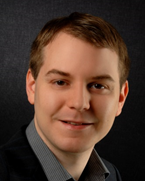}}]
	{Stefan Zernetsch} Stefan Zernetsch received the B.Eng. and the M.Eng. degree in Electrical Engineering and Information Technology from the University of Applied Sciences Aschaffenburg, Germany, in 2012 and 2014, respectively. Currently, he is working on his PhD thesis in cooperation with the Faculty of Electrical Engineering and Computer Science of the University of Kassel, Germany.
\end{IEEEbiography}

\vskip -15mm

\begin{IEEEbiography}[{\includegraphics[width=0.9in,height=1.125in,clip,keepaspectratio]{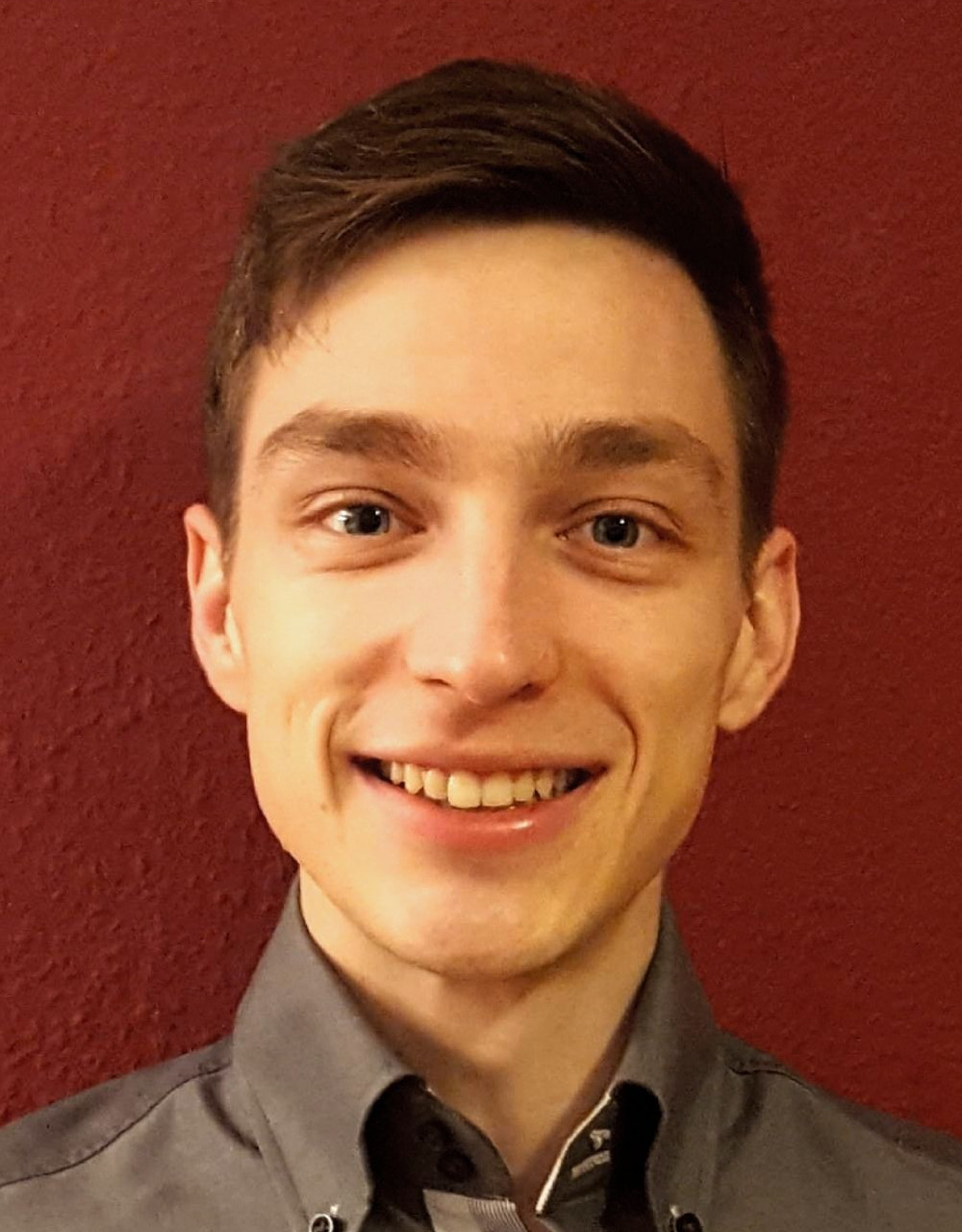}}]
	{Andreas Hubert} Andreas Hubert received the B.Eng. and the M.Eng. degree in Electrical Engineering and Information Technology from the University of Applied Sciences Aschaffenburg, Germany, in 2016 and 2017, respectively. There he conducts research in pattern recognition and short-term behaviour recognition of traffic participants. 
\end{IEEEbiography}

\vskip -15mm

\begin{IEEEbiography}[{\includegraphics[width=0.9in,height=1.15in,clip,keepaspectratio]{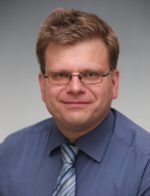}}]
	{Bernhard Sick} Bernhard Sick received the diploma degree in 1992, the PhD degree in 1999, and the ''Habilitation'' degree in 2004, all in computer science, from the University of Passau, Germany. He is currently a full professor of intelligent embedded systems at the Faculty for Electrical Engineering and Computer Science, University of Kassel, Germany. He is an associate editor of the IEEE Transactions on Systems, Man, and Cybernetics-Part B. He holds one patent and received several thesis, best paper, teaching. He is a member of the IEEE (Systems, Man, and Cybernetics Society, Computer Society, and Computational  Intelligence Society) and GI (Gesellschaft fuer Informatik).  
\end{IEEEbiography}

\vskip -12mm

\begin{IEEEbiography}[{\includegraphics[width=0.9in,height=1.125in,clip,keepaspectratio]{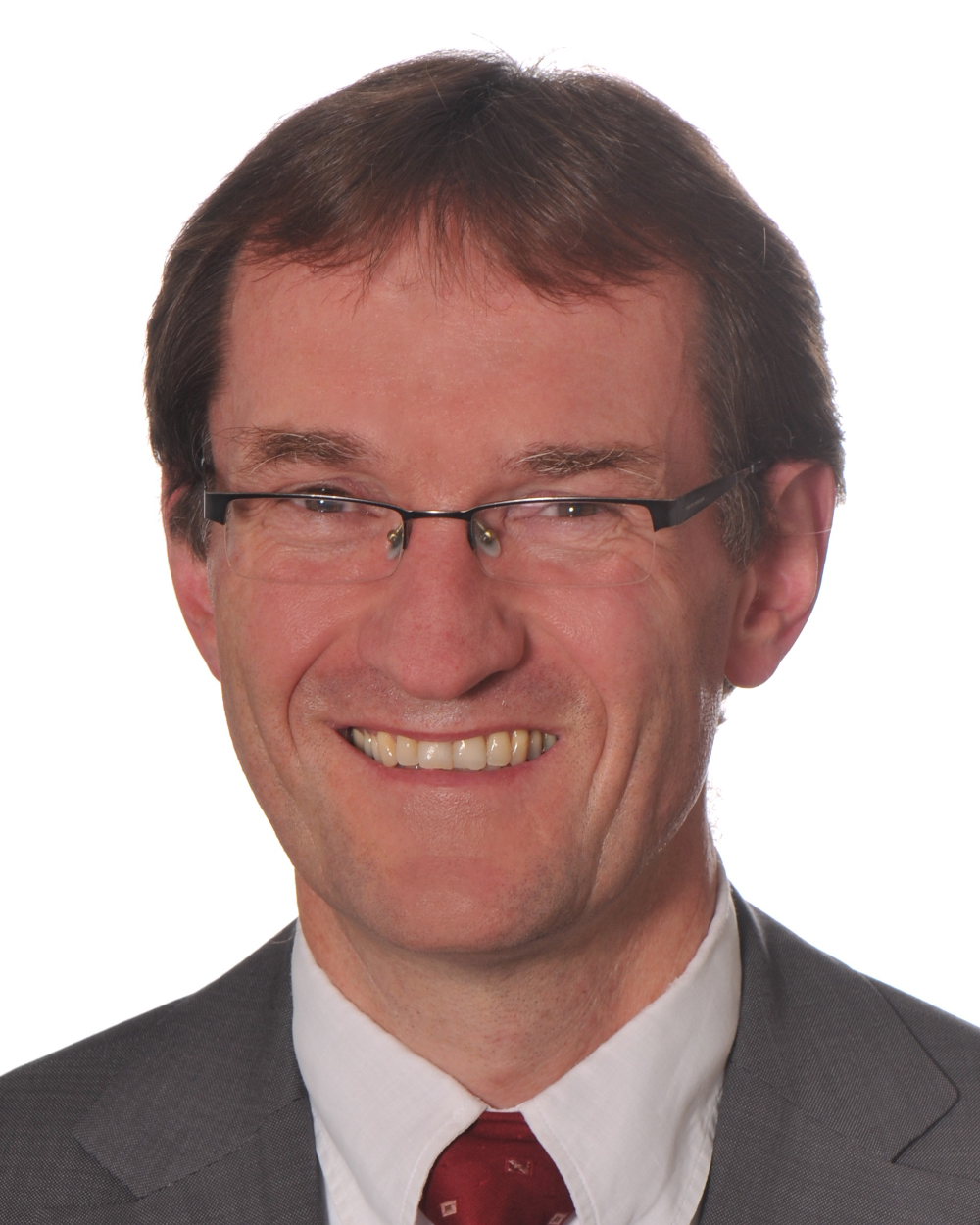}}]
	{Konrad Doll} Konrad Doll received the Diploma (Dipl.-Ing.) degree and the Dr.-Ing.
	degree in Electrical Engineering and Information Technology from the
	Technical University of Munich, Germany, in 1989 and 1994, respectively.
	In 1994 he joined the Semiconductor	Products Sector of Motorola, Inc. (now Freescale Semiconductor,	Inc.). In 1997 he was appointed to professor at the
	University of Applied Sciences Aschaffenburg in the field of
	computer science and digital systems design. Since 2016 he is a research Professor for cooperative automated traffic systems. 
	Konrad Doll is member of the IEEE. 
\end{IEEEbiography}

\end{document}